\def\year{2021}\relax
\documentclass[letterpaper]{article} 
\usepackage{aaai21}  
\usepackage{times}  
\usepackage{helvet} 
\usepackage{courier}  
\usepackage[hyphens]{url}  
\usepackage{graphicx} 
\usepackage{natbib}  
\usepackage{caption} 
\usepackage{amsmath}
\usepackage{booktabs}

\title{Context-Enhanced Entity and Relation Embedding for \\Knowledge Graph Completion (Student Abstract)}
\author {
        Ziyue Qiao,
        Zhiyuan Ning,
        Yi Du\thanks{Corresponding author},
        Yuanchun Zhou\\
}
\affiliations {
    Computer Network Information Center, Chinese Academy of Sciences, Beijing 100190, China\\
    University of Chinese Academy of Sciences, Beijing, China \\
    \{qiaoziyue,ningzhiyuan,duyi,zyc\}@cnic.cn
    }

\begin{document}

\maketitle

\begin{abstract}
Most researches for knowledge graph completion learn representations of entities and relations to predict missing links in incomplete knowledge graphs. However, these methods fail to take full advantage of both the contextual information of entity and relation. Here, we extract contexts of entities and relations from the triplets which they compose. We propose a model named AggrE, which conducts efficient aggregations respectively on entity context and relation context in multi-hops, and learns context-enhanced entity and relation embeddings for knowledge graph completion. The experiment results show that AggrE is competitive to existing models.
\end{abstract}

\section{Introduction}
Knowledge graphs(KGs) store a wealth of knowledge from real world into structured graphs, which consist of collections of triplets, and each triplet $(h, r, t)$ represents that head entity $h$ is related to tail entity $t$ through a relation type $r$.
KGs have play a significant role in AI-related applications such as recommendation systems, question answering, and information retrieval.
However, the coverage of knowledge graphs nowadays is still far from complete and comprehensive,
researchers have proposed a number of knowledge graph completion(KGC) methods to predict the missing links/relations in the incomplete knowledge graph.

Most state-of-the-art methods for KGC are usually based on knowledge graph embeddings, which normally assign an embedding vector to each entity and relation in the continuous embedding space and train the embeddings based on the existing triplets, i.e., to make the scores for observed graph triplets higher than random ones. However, most of them fail to utilize the context/neighborhood of the entity or relation, which may contain rich and valuable information for KGC.

Recently, some researches have proved the significance of contextual information of entity and relation in KGC.
For example, A2N \cite{bansal2019a2n} and RGHAT \cite{zhang2020relational} propose attention-based methods which leverage the contextual information for link prediction by attending to the neighbor entities and lead to more accurate KGC. PathCon \cite{wang2020entity} considers relational context of the head/tail entity and relational paths between head and tail entity in one model, and finds that they are critical to the task of relation prediction.
However, these researches just utilize entity context or relation context, which may lead to information loss.

In this paper, we aim to take full advantage of both the entity context and relation context for enhancing the KGC task. Specifically, different from the neighborhood definition in traditional KG topology, for each element in each triplet, we extract the pair composed of the other two elements as one neighbor in its context.
Then we propose an efficient model, named AggrE, to alternately aggregate the information of entity context and relation context in multi-hops into entity and relation, and learn context-enhanced entity embeddings and relation embeddings. Then we use the learned embeddings to predict the missing relation $r$ given a pair of entities $(h,?,t)$ to complete knowledge graphs.


\begin{figure}[!t]
  \centering
  \includegraphics[width=0.45\textwidth]{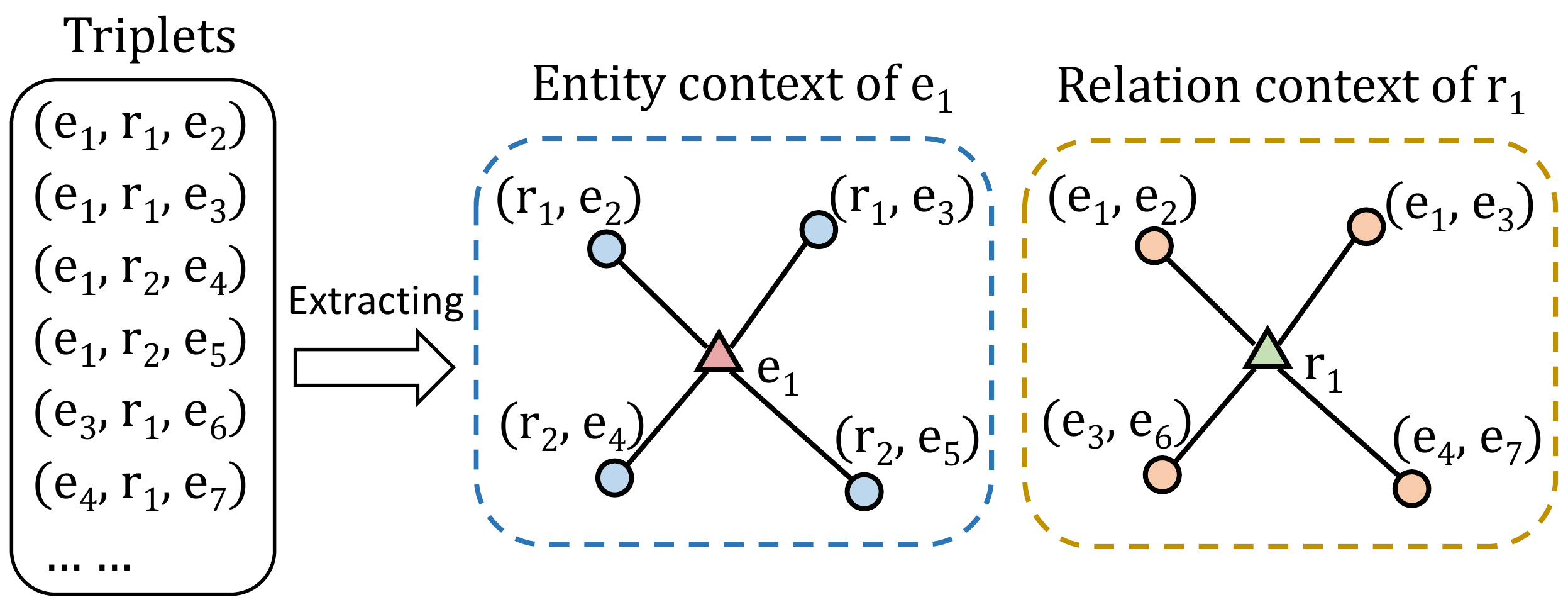}
  \caption{An illustration of extracting entity context and relation context from the triplets of knowledge graphs.}
  \label{context}
\end{figure}

\section{The Proposed Model}
Given a knowledge graph $G=\{(h,r,t)\}\subseteq E \times R \times E$, where $E$ and $R$ are entity set and relation set respectively.
Firstly, we extract contexts of entities and relations from the existing triplets, as shown in Figure \ref{context}.
For an entity $e_i$ in $G$, we define the entity context of $e_i$ as $C_{e_i} = \{(r_j, e_k)|(e_i, r_j, e_k)\in G\vee (e_k, r_j, e_i)\in G\}$, which is actually the set of neighbor entities with their corresponding relations of $e_i$.
For a relations $r_i$ in $G$, we define the relation context of $r_i$ as $C_{r_i} = \{(e_j, e_k)|(e_j, r_i, e_k)\in G\}$, which is actually the two endpoints of $r_i$.
Denote $\mathbf{e}^{(0)}_i$ and $\mathbf{r}^{(0)}_i$ as the randomly initialized embedding of $e_i$ and $r_i$ respectively, our intuition is to aggregate the contextual information
into the representations of each entity and relation to help the prediction.
We define the aggregation functions as:

\begin{align}
&\mathbf{e}^{(l+1)}_i = \mathbf{e}^{(l)}_i + \sum_{(r_j, e_k)\in C_{e_i}}\alpha_{i,j,k}^{(l)}\cdot(\mathbf{r}^{(l)}_j \odot \mathbf{e}^{(l)}_k) \\
&\mathbf{r}^{(l+1)}_i = \mathbf{r}^{(l)}_i + \sum_{(e_j, e_k)\in C_{r_i}}\beta_{i,j,k}^{(l)}\cdot(\mathbf{e}^{(l)}_j \odot \mathbf{e}^{(l)}_k)
\end{align}

where $0\leq l<L$ and $L$ is the number of aggregation layers,
$\mathbf{e}^{(l+1)}_i$ and $\mathbf{r}^{(l+1)}_i$ is the $l$-th layer's output embedding of $e_i$ and $r_i$,
$\odot$ is the element-wise product,
$\alpha_{i,j,k}^{(l)}$ and $\beta_{i,j,k}^{(l)}$ are layer-specific softmax normalization constants,
$\alpha_{i,j,k}^{(l)}$ represent how important each entity context is for $e_i$: $\alpha_{i,j,k}^{(l)} = \frac{\exp(s^{(l)}_{i,j,k})}{\sum\nolimits_{(r_{j'}, e_{k'})\in C_{e_i}} \exp(s^{(l)}_{i,j',k'})}$,
and $\beta_{i,j,k}^{(l)}$ represent how important each relation context is for $r_i$:$\beta_{i,j,k}^{(l)} = \frac{\exp(s^{(l)}_{j,i,k})}{\sum\nolimits_{(e_{j'}, e_{k'})\in C_{r_i}} \exp(s^{(l)}_{j',i,k'})}$.
The $s^{(l)}_{i,j,k}$ represents the score for each possible triplet $(e_i,r_j,e_k)$ after $l$ layers' aggregation, and we use the same score function with DistMult\cite{yang2014embedding} to calculate the triplet scores:

\begin{equation}
s^{(l)}_{i,j,k} = (\mathbf{e}^{(l)}_i)^TDiag(\mathbf{r}^{(l)}_j)\mathbf{e}^{(l)}_k
\end{equation}
Where $Diag(\mathbf{r}^{(l)}_j)$ is a diagonal matrix with $\mathbf{r}^{(l)}_j$ in its diagonal.
After $L$ layers' aggregation, we can obtain the final output $\mathbf{e}^{(L)}_i$ and $\mathbf{r}^{(L)}_i$ of each entity and relation,  which contain neighbor information from their $L$-hops contexts. Then we conduct a softmax loss function on the final triplet scores to compute the likelihood of predicting the correct relations:

\begin{equation}
\mathcal{L} = - \sum_{(e_i, r_j, e_k)\in G} \log \frac{\exp(s^{(L)}_{i,j,k})}{\sum_{r_{j'}\in R}\exp(s^{(L)}_{i,j',k})}
\end{equation}

where $R$ is the set of relations. We use a mini-batch Adam optimizer to minimize $\mathcal{L}$. The difference between our aggregation model and Graph Neural Network(GNN) is that instead of using complex matrix transformation, we use element-wise products to obtain neighborhood information and add it directly to central nodes, as the embedding itself can be regarded as trainable transformation parameters. Also our model is expected to be more efficient and suitable for larger knowledge graph.

%

\section{Experiments}

We conduct experiments on two widely used KG benchmarks: FB15K-237 and WN18RR. Noted that the trainable parameters in our model are only entity and relation embeddings, for a fair comparison, we choose 5 traditional baselines with a small amount of parameters. We use Mean Reciprocal Rank (MRR, the mean of all the reciprocals of predicted ranks), Mean Rank (MR, the mean of all the predicted ranks), and Hit@3(the proportion of correctly predicted entities ranked in the top 3
predictions) as evaluation metrics. In the experiment, we set the embedding dimensionality as 256, the learning rate as 5e-3, l2 penalty coefficient as 1e-7, batch size as 512 and a maximum of 20 epochs. We set the number of aggregation layers $L$ as 2 for WN18RR and 4 for FB15K-237 repectively. 

\begin{table}[]
\scalebox{0.87}{
\begin{tabular}{l|lll|lll}
\toprule
         & \multicolumn{3}{c|}{WN18RR}                                                   & \multicolumn{3}{c}{FB15K-237}                                                \\ \cmidrule{2-7}
         & \multicolumn{1}{c}{MRR} & \multicolumn{1}{c}{MR} & \multicolumn{1}{c|}{Hit@3} & \multicolumn{1}{c}{MRR} & \multicolumn{1}{c}{MR} & \multicolumn{1}{c}{Hit@3} \\ \midrule
TransE   & 0.784                   & 2.079                  & 0.870                      & 0.966                   & 1.352                  & 0.984                     \\
Distmult & 0.847                   & 2.024                  & 0.891                      & 0.875                   & 1.927                  & 0.936                     \\
ComplEx  & 0.840                   & 2.053                  & 0.880                      & 0.924                   & 1.494                  & 0.970                     \\
SimplE   & 0.730                   & 3.259                  & 0.755                      & \textbf{0.971}         & 1.407                  & 0.987                     \\
RotatE   & 0.799                   & 2.284                  & 0.823                      & 0.970                   & 1.315                  & 0.980                     \\ \midrule
AggrE    & \textbf{0.953}   & \textbf{1.136}   & \textbf{0.989}  & 0.966    & \textbf{1.171}  & \textbf{0.989}                     \\ \bottomrule
\end{tabular}}
\caption{Results of relation prediction. The results of baselines are taken from \cite{wang2020entity}}
\label{tab:result}
\end{table}

As shown in Table \ref{tab:result}, The results indicate that AggrE significantly outperforms all the baselines on two benchmarks, which indicates the effectiveness of AggrE. Specifically, the improvement is rather significant for WN18RR, where the links between entities are more sparse than in FB15K-237, this may because without using extra parameters other than embeddings, AggrE is less prone to overfitting. Besides, AggrE achieves substantial improvements against DistMult on all the metrics, and noted that they have the similar objective functions, it indicates that aggregating contextual information for entities and relations is valuable and can great improve the performance of prediction.

\section{Conclusion}
In this paper, we specify a novel definition on the context/neighborhood of entity and relation in KGs, and propose a multi-layer aggregation models to compose contextual information to embeddings for KGC. In the future, we will explore more possible aggregation functions in our model.

\section{Acknowledgments}
This research was supported by the NSFC under Grant 61836013, National Key R\&D Plan of China (2016YFB0501901), Beijing Nova Program of Science and Technology (Z191100001119090).

\bibliography{bibfile}

\begin{thebibliography}{4}
\providecommand{\natexlab}[1]{#1}
\providecommand{\url}[1]{\texttt{#1}}
\providecommand{\urlprefix}{URL }
\expandafter\ifx\csname urlstyle\endcsname\relax
  \providecommand{\doi}[1]{doi:\discretionary{}{}{}#1}\else
  \providecommand{\doi}{doi:\discretionary{}{}{}\begingroup
  \urlstyle{rm}\Url}\fi

\bibitem[{Bansal et~al.(2019)Bansal, Juan, Ravi, and McCallum}]{bansal2019a2n}
Bansal, T.; Juan, D.-C.; Ravi, S.; and McCallum, A. 2019.
\newblock A2N: attending to neighbors for knowledge graph inference.
\newblock In \emph{Proceedings of the 57th Annual Meeting of the Association
  for Computational Linguistics}, 4387--4392.

\bibitem[{Wang, Ren, and Leskovec(2020)}]{wang2020entity}
Wang, H.; Ren, H.; and Leskovec, J. 2020.
\newblock Entity Context and Relational Paths for Knowledge Graph Completion.
\newblock \emph{arXiv preprint arXiv:2002.06757} .

\bibitem[{Yang et~al.(2014)Yang, Yih, He, Gao, and Deng}]{yang2014embedding}
Yang, B.; Yih, W.-t.; He, X.; Gao, J.; and Deng, L. 2014.
\newblock Embedding entities and relations for learning and inference in
  knowledge bases.
\newblock \emph{arXiv preprint arXiv:1412.6575} .

\bibitem[{Zhang et~al.(2020)Zhang, Zhuang, Zhu, Shi, Xiong, and
  He}]{zhang2020relational}
Zhang, Z.; Zhuang, F.; Zhu, H.; Shi, Z.-P.; Xiong, H.; and He, Q. 2020.
\newblock Relational Graph Neural Network with Hierarchical Attention for
  Knowledge Graph Completion.
\newblock In \emph{AAAI}, 9612--9619.

\end{thebibliography}

\end{document}